\documentclass{article}

\usepackage[nonatbib,preprint]{nips_2018}

\usepackage[utf8]{inputenc} 
\usepackage[T1]{fontenc}    
\usepackage{hyperref}       
\usepackage{url}            
\usepackage{booktabs}       
\usepackage{amsfonts}       
\usepackage{nicefrac}       
\usepackage{microtype}      
\usepackage{subfig}
\usepackage{graphicx}
\usepackage{multirow}
\usepackage{calc}
\usepackage{multicol}
\usepackage{threeparttable}

\usepackage{array}
\usepackage[cmex10]{amsmath}
\usepackage[T1]{fontenc}
\usepackage{xspace}
\newcommand{\PolyNeuron}{\textup{\textrm{PolyNeuron}}\xspace}
\newcommand{\PolyNeuronRelaxed}{\textup{\textrm{PolyNeuron-R}}\xspace}

\title{PolyNeuron: Automatic Neuron Discovery via Learned Polyharmonic Spline Activations}

\author{Andrew Hryniowski$^{1,2,3}$, Alexander Wong$^{1,2,3}$\\
			$^{1}$ Video and Image Processing Lab, Systems Design Engineering, University of Waterloo\\
			$^{2}$ Waterloo Artificial Intelligence Institute, Waterloo, ON\\
			$^{3}$ DarwinAI Corp., Waterloo, ON\\ 			
			\texttt{$\{$apphryni, a28wong$\}$@uwaterloo.ca}
	}

\begin{document}
\maketitle

\begin{abstract}

Automated deep neural network architecture design has received a significant amount of recent attention. However, this attention has not been equally shared by one of the fundamental building blocks of a deep neural network, the neurons. In this study, we propose PolyNeuron, a novel automatic neuron discovery approach based on learned polyharmonic spline activations. More specifically, PolyNeuron revolves around learning polyharmonic splines, characterized by a set of control points, that represent the activation functions of the neurons in a deep neural network. A relaxed variant of PolyNeuron, which we term PolyNeuron-R, loosens the constraints imposed by PolyNeuron to reduce the computational complexity for discovering the neuron activation functions in an automated manner. Experiments show both PolyNeuron and PolyNeuron-R lead to networks that have improved or comparable performance on multiple network architectures (LeNet-5 and ResNet-20) using different datasets (MNIST and CIFAR10). As such, automatic neuron discovery approaches such as PolyNeuron is a worthy direction to explore.

\end{abstract}

\section{Introduction}
\label{sec:label}

Designing a deep neural network (DNN) can be a time consuming process due to the number of possible configurations and the time required to train each configuration. One must not only decide on the type of architecture (e.g., convolutional neural networks (CNNs)~\cite{lecun1998gradient}, recurrent neural networks (RNNs)~\cite{hochreiter1997long}, etc.), but the detailed architecture design (e.g., number of layers, the type of layers to use, number of neurons per layer, type of activation function to use, etc.). To tackle the bottleneck in network architecture design, recent research efforts have focused on automating the search for optimal network architectures~\cite{zoph2016neural, xie2017genetic, shafiee2018deep, stanley2002evolving,wong2018}.

Unlike automated architecture design, automated neuron design methodologies have not received the same amount of research focus. A recent approach to neuron design is to use an intelligent brute force search scheme in which designs are iteratively tested and improved upon~\cite{ramachandran2018searching}. Such a technique is computationally expensive and not tractable for many practitioners. Other research in this area has focused on extending classic neuron designs by parameterizing the neurons with scaling parameters~\cite{he2015delving}, or by using ensemble methods~\cite{harmon2017activation, sutfeld2018adaptive}. These methods are either static or restricted to limited change. To expand the scope of possible learnable neuron designs, piece-wise functions have also been explored~\cite{jin2016deep, agostinelli2014learning, douzette2017b, scardapane2017learning}.

In this study, we propose PolyNeuron, a novel automatic neuron discovery approach based on learned polyharmonic spline activations, along with a relaxed variant called PolyNeuron-R.  The remainder of this paper is organized as follows. Section~\ref{sec:method} formalizes the proposed PolyNeuron and PolyNeuron-R approaches, and Section~\ref{sec:exp_results} compares them with existing methods and discusses design properties of the proposed approaches.

\section{Methodology}
\label{sec:method}

Parametric neuron design discovery approaches leverage tunable weights within the neuron design to provide a deep neural network the potential to learn a more optimal activation function for any given neuron.  Towards that goal, we first introduce a new neuron design discovery approach named \PolyNeuron, which is based on learning polyharmonic spline activations. We then introduce a second neuron design discovery approach named \PolyNeuron-R, with the goal of reducing the computational requirements of \PolyNeuron.

\subsection{PolyNeuron: Polyharmonic Spline Activations}

Existing parametric neuron discovery approaches are either linear in nature or require a closest control-point search to select the appropriate regional parameters. In addition, these parametric approaches often require regularization constraints to remain stable during training. To tackle these problems, the proposed \PolyNeuron approach leverages  polyharmonic spline based activation functions. Polyharmonic splines are useful tools for interpolation as they do not require a control point selection step and have built-in constraints. For some neuron activation function $a$, in a DNN, with $s$ control points, let \PolyNeuron be defined as,
\begin{equation}
\label{eq:interp_equation}
    \PolyNeuron_a(x_a) = y_a =\sum_{i=0}^{s-1} w_{ai} U_k(\vert x_a - c_{axi} \vert) + v_{a0} x_a + v_{a1}
\end{equation}
where $x_a$ and $y_a$ are the input and output for the $a^{th}$ neuron activation function, respectively. $U_k(\cdot)$ is the $k^{th}$ polyharmonic radial basis function (RBF), and $c_{axi}$ is the $x$ coordinate of the $i^{th}$ control point. Interestingly, the form of Equation~\ref{eq:interp_equation} resembles that of a radial basis function network~\cite{broomhead1988radial}, but with the addition of the linear component $v_{a0}x_a + v_{a1}$. For a set of $s$ trainble control points $\{(c_{axi}, c_{ayi})\}_s$, where $c_{ayi}$ is the $y$ coordinate of the $i^{th}$ control point, the polyharmonic spline parameters $\{w_{ai}\}_s$, $v_{a0}$, and $v_{a1}$ can be calculated by solving the following linear system,
\begin{equation}
\label{eq:spline_matrix}	
	\begin{bmatrix}
		U(\vert \vec{c}_{ax} - \vec{c}^T_{ax} \vert)  & \textbf{C}_a \\
		\textbf{C}_a^T & \textbf{0}
	\end{bmatrix}
	\cdot
	\begin{bmatrix}
		\vec{w_a} \\
		\vec{v_a}
	\end{bmatrix}
	=
	\begin{bmatrix}
	\vec{c_{ay}} \\
	\vec{0}
	\end{bmatrix}
\end{equation}
where  $\textbf{C}_a = \begin{bmatrix} \vec{c}_{ax} & \textbf{1} \end{bmatrix}$, $\vec{c}_{ax} = [c_{ax0} \dotsc c_{ax,s-1}]^T$, $\vec{c}_{ay} = [c_{ay0} \dotsc c_{ay,s-1}]^T$, $\vec{w}_a = [w_{a0} \dotsc w_{a,s-1}]^T$, and $\vec{v_a} = [v_{a0}\,v_{a1}]^T$.  Finally, the polyharmonic radial basis function $U_k$ is defined as,
\begin{equation}
\label{eq:radial_basis_function}	
	U_k(r) =
	\begin{cases}
	r^k, & \text{if $k$ is odd} \\
	r^k \log{r}, & \text{if $k$ is even}
	\end{cases}
\end{equation}
Note for all $k$'s $U_k(0) = 0$.

For every \PolyNeuron there are two considerations to be taken into account: i) the number of control points $s$, and ii) the polyharmonic RBF $U_k$. The number of control points in \PolyNeuron causes the square matrix in Equation~\ref{eq:spline_matrix} to grow in a quadratic manner since each control point must be \textit{compared} to every other control point in the spline. As a result, the number of operations to solve the linear system grows cubically with respect to $s$. The RBF number $k$ causes \PolyNeuron's computation complexity to grow linearly with respect to $s^2$ (the approximately number of control point comparisons in Equation~\ref{eq:spline_matrix}).

\subsection{PolyNeuron-R: Relaxed PolyNeuron}

\PolyNeuron requires the linear system in Equation~\ref{eq:spline_matrix} to be solved each time the control points for a given neuron activation function are updated during backpropagation. In addition, the backpropagation algorithm must calculate the gradient back through the linear system solution to update the control points. Both these facts result in a significant increase to the number of operations required for feature activation learning. To reduce the computational complexity of \PolyNeuron, let us take a look at the three strict constraints embedded in Equation~\ref{eq:spline_matrix},

\begin{tabular}{m{0.03\linewidth} l m{0.3\linewidth} l m{0.2\linewidth} l m{0.2\linewidth}}

    &
    1. & $\PolyNeuron_a(c_{axi})=c_{ayi}$ &
    2. & $\sum_{i=0}^{s-1} w_{ai} c_{axi} = 0$ &
    3. & $\sum_{i=0}^{s-1} w_{ai} = 0$
\end{tabular}

We propose \PolyNeuron-R, an approximation of \PolyNeuron that modifies the above constraints in order to remove solving the linear system during each backpropagation iteration. Two adjustments are made to \PolyNeuron: first, the parameters $\vec{w}_a$, $v_{a0}$, and $v_{a1}$ will be directly learned instead of solving for them using Equation~\ref{eq:spline_matrix}, and second, the product constraint $\sum_{i=0}^{s-1} w_{ai} c_{axi} = 0$ and the sum constraint $\sum_{i=0}^{s-1} w_{ai} = 0$ are added as regularization terms to the model's loss function,
\begin{equation}
    L_{\PolyNeuron-R} = L_{model} + \lambda_{prod} \frac{1}{A}\sum_a^A | \sum_i^{s-1} w_{ai} c_{axi} | + \lambda_{sum}\frac{1}{A} \sum_a^A | \sum_i^{s-1} w_{ai} |
\end{equation}
where $L_{\PolyNeuron-R}$ is the loss function for a network using \PolyNeuron-R, $L_{model}$ is the loss function for a given model without the adjusted activation function, $A$ is the number of activation functions in a DNN, and $\lambda_{prod}$ and $\lambda_{sum}$ are the regularization strengths for the product constraint and the sum constraint, respectively. By removing the linear system solution calculation, \PolyNeuron-R reduces the number of computations required for activation compared to \PolyNeuron since $s$ now has a linearly effect on the computational complexity.

\section{Experimental Results}
\label{sec:exp_results}

In this section, the proposed \PolyNeuron and \PolyNeuron-R are tested on two different benchmarks: LeNet-5~\cite{lecun1998gradient} using MNIST~\cite{lecun1998mnist}, and ResNet-20~\cite{he2016deep} using CIFAR-10~\cite{krizhevsky2009learning}. For each benchmark, \PolyNeuron and \PolyNeuron-R are compared to the well-known static activation function ReLU, and two learned parametric activation functions (Swish~\cite{ramachandran2018searching} and Adaptive Piecewise Linear (APL)~\cite{agostinelli2014learning}).

The various design decisions and training parameters for \PolyNeuron and \PolyNeuron-R are as follows. The LeNet-5 benchmark trains for 30 epochs with an initial learning rate of $10^{-3}$. The learning rate is reduced by an order or magnitude at 20 epochs and 25 epochs. The ResNet-20 benchmark trains for 500 epochs with an initial learning rate of $10^{-3}$. The learning rate is reduced by an order or magnitude at 250 epochs and 375 epochs. For both benchmarks, the networks use the Adam optimizer~\cite{kingma2014adam} for backpropagation, using random weight initialization, and a $L_2$ weight decay rate of $10^{-4}$. The weight decay in each network does not effect the parameters for the parametric activation functions. The training images in each dataset are padded by 4 pixels on all sizes and randomly cropped to the original input dimensions, then randomly flipped along the vertical axis, and finally normalized. A batch size of 128 is used for each network. Remember that \PolyNeuron learns a set of $s$ control points $\{(c_{axi}, c_{ayi})\}_s$, and \PolyNeuron-R learns a set of $s$ $x$-axis control points $\{c_{axi}\}_s$, a set of $s$ control point weights $\{w_{ai}\}_s$, an input scaling factor $v_{a0}$, and constant offset $v_{a1}$. Table~\ref{tab:experiment_results} shows the mean validation classification error on the two benchmarks across 5 training runs.

\newcolumntype{N}{>{\centering\arraybackslash}m{0.15\linewidth}}
\newcolumntype{G}{>{\bfseries\centering\arraybackslash}m{0.15\linewidth}}
\renewcommand{\arraystretch}{1.1}
\captionsetup[table]{skip=10pt}

\begin{table}[!h]
    \caption{Comparison between tested activation functions ($\% error$)}
    \centering
    \begin{threeparttable}
    \begin{tabular}{lNN}
        \toprule
        \multicolumn{1}{l}{\textbf{Activation Function}} &
        \multicolumn{1}{G}{LeNet-5 MNIST} &
        \multicolumn{1}{G}{ResNet-20 CIFAR-10}
        \\ [0.5ex]
        \hline
        \multicolumn{1}{l}{ReLU}
        & 2.03 & 9.31  \\ [0.5ex]
        \hline
        \multicolumn{1}{l}{Swish~\cite{ramachandran2018searching}}
        & 1.91 & \textbf{8.60}  \\ [0.5ex]
        \multicolumn{1}{l}{APL~\cite{agostinelli2014learning}}
        & 1.80 & 9.09  \\ [0.5ex]
        \hline
        \multicolumn{1}{l}{\PolyNeuron$^*$ (ours)}
        & 1.75 & 8.68  \\ [0.5ex]
        \multicolumn{1}{l}{\PolyNeuron-R$^*$(ours)}
        & \textbf{1.66} & 8.64   \\ [0.5ex]
        \bottomrule
    \end{tabular}
    \begin{tablenotes}
        \small
        \item $^* s=3, k=3$
    \end{tablenotes}
    \end{threeparttable}
    \label{tab:experiment_results}
\end{table}

Multiple value combinations for $s$ and $k$ for \PolyNeuron and \PolyNeuron-R were tested. Experimentally, even values of $k$ (i.e., $U_k$ contains $log(\cdot)$) reduce the performance of each benchmark for \PolyNeuron. No performance gains for $k > 3$ were realized and only served to increase the number of required computations. Moderate values of $s$ result in significantly longer training times for networks using \PolyNeuron and did not increase validation performance. For \PolyNeuron, the best choices for the number of control points and the RBF are $s=3$ and $k=3$, respectively, as they restrict the number computations without a reduction in performance. To remain consistent during comparisons, $s=3$ and $k=3$ were also used for \PolyNeuron-R.

\PolyNeuron and \PolyNeuron-R with random weight initializations tend to be numerically unstable during early training. For example, if a boundary control point (i.e., a control point with only one neighbour) is close enough to the neighbouring control point, the function beyond the boundary control point can quickly grow towards $\pm\infty$. A \textit{ReLU-like} initialization is used for both \PolyNeuron and \PolyNeuronRelaxed to prevent this stability issue. For \PolyNeuron $\{(c_{axi}, c_{ayi})\}_3 = \{(-1,0), (0,0), (1,1)\}$, and for \PolyNeuronRelaxed the parameters $\{w_{ai}\}_3$, $v_{a0}$, and $v_{a1}$ are set such that the points $\{(-1,0),(0,0),(1,1)\}$ lie on the activation curve.

\begin{figure}[!t]
    \centering
    \begin{tabular}{c}
        \subfloat[\PolyNeuron Example\label{subfig:pn_sample}]{\includegraphics[width=1.0\linewidth]{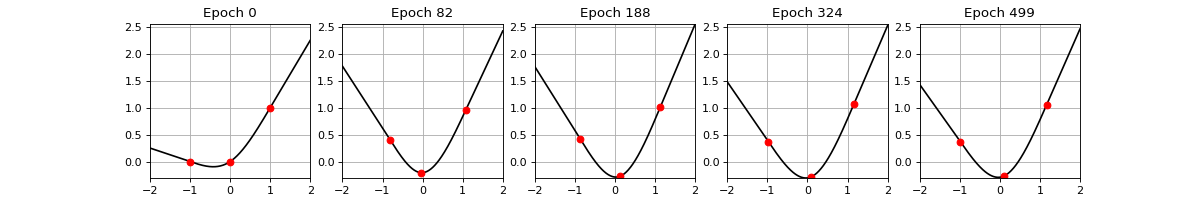}} \\

        \subfloat[\PolyNeuron-R Example\label{subfig:pnr_sample}]{\includegraphics[width=1.0\linewidth]{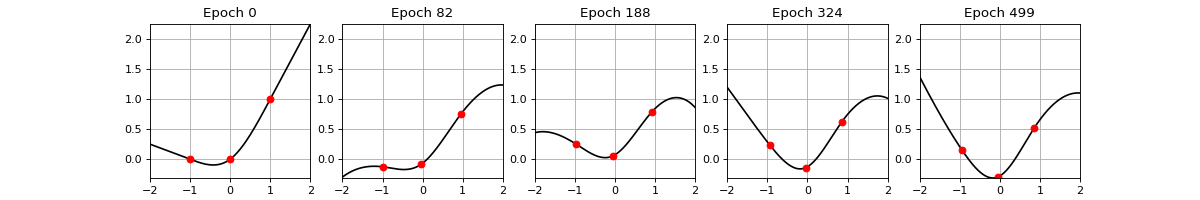}} \\

    \end{tabular}

    \caption{\small An example of \PolyNeuron~\protect\subref{subfig:pn_sample} and \PolyNeuron-R~\protect\subref{subfig:pnr_sample} evolving over epochs during separate training runs on the ResNet-20 CIFAR-10 benchmark. The red dots in each figure indicate the control point locations. \PolyNeuron remains approximately linear at the boundaries while \PolyNeuron-R does not. The evolution of \PolyNeuron and \PolyNeuron-R across epochs is similar in that the majority of changes to both activation functions happen earlier on in training and the control point locations are relatively consistent in the $x$-axis. }
    \vspace{-0.2in}
    \label{fig:spline_examples}
\end{figure}

Multiple values for $\lambda_{prod}$ and $\lambda_{sum}$ are tested for \PolyNeuronRelaxed with $s=3$ and $k=3$. Experimentally, $\lambda_{prod}$ provides no performance benefits and in some cases hinders test performance. This is most likely because the control point values $\{c_{axi}\}_3$ do not deviate far from initialization without $\lambda_{prod}$ and tend towards $0$ with $\lambda_{prod}$. As the points $\{c_{axi}\}_3$ approach each other the activation curve is more sensitive to updates affecting beyond the boundary control points. Thus, a $\lambda_{prod}$ is set to $0$. A value of $10^{-2}$ is found to be suitable for $\lambda_{sum}$. Setting both $\lambda_{prod}$ and $\lambda_{sum}$ to $0$ causes the training performance across initializations to become inconsistent.  Figure~\ref{fig:spline_examples} visualizes examples of \PolyNeuron and \PolyNeuron-R evolving over epochs throughout training on the ResNet-20 CIFAR-10 benchmark.

It can be observed that \PolyNeuronRelaxed had the best average performance for the LeNet-5 benchmark, and Swish (note that this refers to the learned variant of Swish) had the best average performance for the ResNet-20 CIFAR-10 benchmark. \PolyNeuronRelaxed and APL are more sensitive to higher learning rates then the other activation functions; that is, they converge to a lower validation accuracy prior to decreasing the learning rate for the first time. Reducing the initial learning rate for these networks may result in improved final performance. Despite the reduced constraints and reduced computational requirements, \PolyNeuronRelaxed outperformed \PolyNeuron on both benchmarks. The performance difference is surprising as the choices for $s$ and $k$ were motivated by \PolyNeuron and not \PolyNeuronRelaxed. It is likely that the strict constraints imposed by Equation~\ref{eq:spline_matrix} prevents the \PolyNeuron parameters from converging to more optimal values.

The above comparison has demonstrated \PolyNeuron is a viable neuron design discovery approach. However, \PolyNeuronRelaxed's reduced computational cost and superior performance makes it a more attractive candidate. \PolyNeuronRelaxed was only tested using a single set of parameters ($s=3$ and $k=3$) and will require additional comparison to fully realize its potential. The performance of the two proposed neuron design discovery approaches indicated that there is potential room for improving automated neuron design procedures, thus making it an attractive candidate for future research.

\subsubsection*{Acknowledgments}
We would like to thank Natural Sciences and Engineering Research Council (NSERC) and Canada Research Chairs program.

\medskip
\small

\bibliography{and}
\bibliographystyle{ieeetr}

\end{document}